\documentclass{article} 

\usepackage[usenames,dvipsnames]{xcolor}
\usepackage{soul}
\sethlcolor{SkyBlue}

\usepackage{iclr2018_conference,times}
\usepackage{hyperref}
\usepackage{url}

\usepackage{booktabs,multirow,makecell}       
\usepackage{amsfonts}       
\usepackage{nicefrac}       
\usepackage{microtype}      

\usepackage{graphicx}
\usepackage{algorithm}

\usepackage[noend]{algpseudocode}
\usepackage{subcaption}
\usepackage[export]{adjustbox}
\usepackage{caption}

\usepackage{amsmath}
\usepackage{amsthm}
\usepackage{amsfonts}
\usepackage{amssymb}

\newtheoremstyle{normal}
  {.2\baselineskip\@plus.05\baselineskip\@minus.05\baselineskip}
  {.05\baselineskip\@plus.05\baselineskip\@minus.05\baselineskip}
  {\itshape}
  {}
  {\bfseries}
  {.}
  { }
  {}

\newtheoremstyle{cited}%
  {.2\baselineskip\@plus.05\baselineskip\@minus.05\baselineskip}
  {.05\baselineskip\@plus.05\baselineskip\@minus.05\baselineskip}
  {\itshape}
  {}
  {\bfseries}
  {.}
  {.5em}
  {\thmname{#1} \thmnumber{#2} \thmnote{\normalfont#3}}

\theoremstyle{normal}

\newtheorem{define}{Definition}
\newtheorem{prop}{Proposition}

\theoremstyle{cited}

\newcommand{\alg}[1]{\textnormal{\textsc{#1}}}

\newcommand{\R}{\mathbb{R}}

\newcommand{\mb}[1]{\mathbf{#1}}
\newcommand{\mc}[1]{\mathcal{#1}}

\newcommand{\w}{\mathbf{w}}
\newcommand{\x}{\mathbf{x}}

\newcommand{\z}{\mathbf{z}}

\usepackage{etoolbox}
\makeatletter
\newcount\c@additionalboxlevel
\newcount\c@maxboxlevel
\patchcmd\@combinedblfloats{\box\@outputbox}{%
  \stepcounter{additionalboxlevel}%
  \box\@outputbox
}{}{\errmessage{\noexpand\@combinedblfloats could not be patched}}

\AtBeginShipout{%
  \ifnum\value{additionalboxlevel}>\value{maxboxlevel}%
    \typeout{Warning: maxboxlevel might be too small, increase to %
      \the\value{additionalboxlevel}%
    }%
  \fi 
  \@whilenum\value{additionalboxlevel}<\value{maxboxlevel}\do{%
    \typeout{* Additional boxing of page `\thepage'}%
    \setbox\AtBeginShipoutBox=\hbox{\copy\AtBeginShipoutBox}%
    \stepcounter{additionalboxlevel}%
  }%
  \setcounter{additionalboxlevel}{0}%
}
\makeatother

\newcommand{\ds}[1]{{(#1)}}
\newcommand{\h}{\mb{h}}
\newcommand{\algname}{\alg{ftprop}}
\newcommand{\algnamemb}{\algname\alg{-mb}}

\DeclareMathOperator*{\sign}{sign}
\DeclareMathOperator*{\sathinge}{sat\_hinge}
\DeclareMathOperator*{\softhinge}{soft\_hinge}
\DeclareMathOperator*{\satrelu}{sat\_ReLU}
\DeclareMathOperator*{\qrelu}{qReLU}
\DeclareMathOperator*{\step}{step}

\usepackage[compact]{titlesec}
\title{Deep Learning as a Mixed Convex-\\Combinatorial Optimization Problem}

\author{Abram L.~Friesen and Pedro Domingos \\
Paul G. Allen School of Computer Science and Engineering\\
University of Washington\\
Seattle, WA 98195, USA \\
\texttt{\{afriesen,pedrod\}@cs.washington.edu} \\
}

\iclrfinalcopy 

\begin{document}


\maketitle

\begin{abstract}



As neural networks grow deeper and wider, learning networks 
with hard-threshold activations is becoming increasingly important, 
both for network quantization, which can drastically reduce time and 
energy requirements, and for creating large integrated systems of 
deep networks, which may have non-differentiable components and 
must avoid vanishing and exploding gradients for effective learning. 
However, since gradient descent is not applicable to hard-threshold 
functions, it is not clear how to learn networks of them 
in a principled way. We 
address this problem by observing that setting targets for hard-threshold 
hidden units in order to minimize loss is a discrete optimization 
problem, and can be solved as such. 
The discrete optimization goal 
is to find a set of targets such that each unit, including the output, 
has a linearly separable problem to solve. Given these targets, the 
network decomposes into individual perceptrons, which can then be 
learned with standard convex approaches. 
Based on this, we develop a recursive 
mini-batch algorithm for learning deep hard-threshold networks that 
includes the popular but poorly justified straight-through estimator as 
a special case. 
Empirically, we show that our algorithm improves classification accuracy
in a number of settings, including for AlexNet and ResNet-18 
on ImageNet, when compared to the straight-through estimator. \looseness=-1

\end{abstract}

\vspace{-1mm}
\section{Introduction}
\vspace{-1mm}

The original approach to neural classification was to learn 
single-layer models with hard-threshold activations, like the perceptron~\citep{Rosenblatt1958}.
However, it proved difficult to extend these methods to multiple layers, because
hard-threshold units, having zero derivative almost everywhere and being
discontinuous at the origin, cannot be trained by gradient descent.
Instead, the community turned to multilayer networks with soft activation
functions, such as the sigmoid and, more recently, the ReLU, for 
which gradients can be computed efficiently by backpropagation~\citep{Rumelhart1986}.
\looseness=-1

This approach has enjoyed remarkable success, enabling
researchers to train networks with hundreds of layers and learn models that have
significantly higher accuracy on a variety of tasks than any previous approach.
However, as networks become deeper and wider, there has been a growing
trend towards using hard-threshold activations for quantization purposes,
where they enable binary or low-precision 
inference (e.g.,~\citet{Hubara2016a,Rastegari2016,Zhou2016,Lin2016a,Zhu2017}) 
and training (e.g.,~\citet{Lin2016b,Li2017,Tang2017,Micikevicius2017}), which can greatly
reduce the energy and computation time required by modern deep networks.
Beyond quantization, the scale of the output of hard-threshold units
is independent of (or insensitive to) the scale of their input,
which can alleviate vanishing and exploding gradient issues
and should help avoid some of the pathologies that occur during 
low-precision training with backpropagation~\citep{Li2017}.
Avoiding these issues is crucial for developing large systems of deep networks 
that can be used to perform even more complex tasks.


For these reasons, we are interested in developing well-motivated and efficient techniques
for learning deep neural networks with hard-threshold units.
In this work, we propose a framework for learning deep hard-threshold networks that 
stems from the observation that hard-threshold units output
discrete values, indicating that combinatorial optimization may provide a principled
method for training these networks. By specifying a set of discrete targets
for each hidden-layer activation, the network decomposes into many individual perceptrons,
each of which can be trained easily given its inputs and targets. The difficulty in
learning a deep hard-threshold network is thus in setting the targets so that
each trained perceptron -- including the output units -- 
has a linearly separable problem to solve and thus can achieve its targets.
We show that networks in which this is possible can be learned using our
mixed convex-combinatorial optimization framework.

Building on this framework, we then develop a recursive algorithm, 
feasible target propagation (\algname{}), for learning deep hard-threshold networks.
Since this is a discrete optimization problem, we develop heuristics for
setting the targets based on per-layer loss functions.
The mini-batch version of \algname{} can be used to explain and justify the
oft-used straight-through estimator~\citep{Hinton2012,Bengio2013}, 
which can now be seen as an instance of \algname{} with a specific choice
of per-layer loss function and target heuristic.
Finally, 
we develop a novel per-layer loss function that improves learning of
deep hard-threshold networks.
Empirically, we show improvements for our algorithm over the straight-through estimator on 
CIFAR-10 for two convolutional networks and on ImageNet 
for AlexNet and ResNet-18,
with multiple types of hard-threshold activation. 


\vspace{-1mm}
\subsubsection*{Related work}
\vspace{-1mm}

The most common method for learning deep hard-threshold 
networks is to use backpropagation 
with the straight-through estimator (STE)~\citep{Hinton2012,Bengio2013}, 
which simply replaces the derivative of each hard-threshold unit 
with the identity function. 
The STE is used in the quantized network literature (see citations above) 
to propagate gradients through quantized activations, and is 
used in \citet{Shalev-Shwartz2017} for training with flat activations.
Later work generalized the STE to replace the hard-threshold derivative with
other functions, including saturated versions of the 
identity function~\citep{Hubara2016a}.
However, while the STE tends to work quite well in practice, 
we know of no rigorous justification or
analysis of why it works or how to choose replacement derivatives. 
Beyond being unsatisfying in this regard, 
the STE is not well understood and can lead to gradient mismatch errors, 
which compound
as the number of layers increases~\citep{Lin2016a}. 
We show here that the STE, saturated STE, and 
all types of STE that we have seen
are special cases of our framework, thus 
providing a principled justification for it and a basis 
for exploring and understanding alternatives.
\looseness=-1

Another common approach to training with hard-threshold units is to use randomness,
either via stochastic neurons 
(e.g.,~\citet{Bengio2013,Hubara2016a}) 
or probabilistic training methods, such as those of \citet{Soudry2014} or \citet{Williams1992},
both of which are methods for softening hard-threshold units.
In contrast, our goal is to learn networks with deterministic hard-threshold units.

Finally, target propagation (TP)~\citep{LeCun1986,LeCun1987,Carreira-Perpinan2012a,
Bengio2014,Lee2015,Taylor2016a} 
is a method that explicitly associates a target with the 
output of each activation in the network, 
and then updates each layer's weights to make its 
activations more similar to the targets. 
Our framework can be viewed as an instance of TP 
that uses combinatorial optimization to
set discrete targets, whereas previous 
approaches employed continuous optimization to set continuous targets.
The MADALINE Rule II 
algorithm~\citep{Winter1988} can also be seen as a 
special case of our framework and of TP, 
where only one target is set at a time.

\vspace{-1mm}
\section{Learning deep networks with hard-threshold units}
\label{sec:tp}
\vspace{-1mm}



Given a dataset $\mc{D} = \{(\x^\ds{i}, t^\ds{i})\}_{i=1}^m$ with 
vector-valued inputs $\x^\ds{i} \in \R^{n}$ and binary targets
$t \in \{-1, +1\}$, we are interested in learning an 
$\ell$-layered deep neural network with hard-threshold units \looseness=-1

\vspace{-1.25em}
\begin{align}
y = f(\x ; W) = g(W_\ell ~g(W_{\ell-1} \dots g(W_1 \x) \dots)),
\end{align}
\vspace{-1.25em}

with weight matrices $W = \{ W_d : W_d \in \R^{n_{d} \times n_{d-1}} \}_{d=1}^\ell$
and element-wise activation function $g(\x) = \sign(\x)$, 
where $\sign$ is the sign function such that $\sign(x) = 1$ if $x > 0$ and $-1$ otherwise.
Each layer $d$ has $n_d$ units, where we define $n_0 = n$ for the input layer,
and we let $\h_d = g(W_d \dots g(W_1 \x) \dots)$ denote the output of each hidden layer,
where $\h_d = (h_{d1}, \dots, h_{dn_d})$ and $h_{dj} \in \{-1, +1\}$ for each layer $d$ and each unit $j$.
Similarly, we let $\z_d = W_d ~g( \dots g(W_1 \x) \dots )$ denote the pre-activation output of layer $d$.
For compactness, we have incorporated the bias term into the weight matrices.
We denote a row or column of a matrix $W_d$ as $W_{d,:j}$ and $W_{d,j:}$, respectively, 
and the entry in the $j$th row and $k$th column as $W_{d,jk}$.
Using matrix notation, we can write this model as
$Y = f(X; W) = g(W_\ell \dots g(W_1 X) \dots)$, where $X$ is the $n \times m$ matrix of dataset instances and 
$Y$ is the $n_\ell \times m$ matrix of outputs. We let $T_\ell$ denote the matrix of final-layer targets,
$H_d$ denote the $n_d \times m$ matrix of hidden activations at layer $d$, 
and $Z_d$ denote the $n_d \times m$ matrix of pre-activations at layer $d$.
Our goal will be to learn $f$ by finding the weights $W$ that minimize an aggregate loss 
$L(Y, T_\ell) = \sum_{i=1}^m L(y^\ds{i}, t^\ds{i})$ for some convex per-instance loss $L(y, t)$.

In the simplest case, a hard-threshold network with no hidden layers 
is a perceptron $Y = g(W_1 X)$, as introduced by~\citet{Rosenblatt1958}. 
The goal of learning a perceptron, or any hard-threshold network,
is to classify unseen data. 
A useful first step 
is to be able to correctly classify the training data, which we focus on here
for simplicity when developing our framework; 
however, standard generalization techniques such as regularization
are easily incorporated into this framework and we do this for the experiments. 
%
%
%
Since a perceptron is a linear classifier, it is only able to separate 
a linearly-separable dataset.
\looseness=-1

\begin{define}
A dataset $\{(\x^\ds{i}, t^\ds{i} ) \}_{i=1}^m$ 
is \emph{linearly separable} iff there exists a vector 
$\w \in \R^n$ and a real number $\gamma > 0$ 
such that $(\w \cdot \x^\ds{i}) t^\ds{i} > \gamma$ for all $i = 1 \dots m$.
\end{define}

When a dataset is linearly separable, the perceptron 
algorithm is guaranteed to find its separating hyperplane
in a finite number of steps~\citep{Novikoff1962}, where the number of steps
required is dependent on the size of the margin $\gamma$.
However, linear separability is a very strong condition, and
even simple functions, such as XOR, 
are not linearly separable and thus cannot be learned by a perceptron~\citep{Minsky1969}.
We would thus like to be able to learn multilayer hard-threshold networks.\looseness=-1

\begin{figure}[tb]
\begin{centering}
\includegraphics[width=\textwidth]{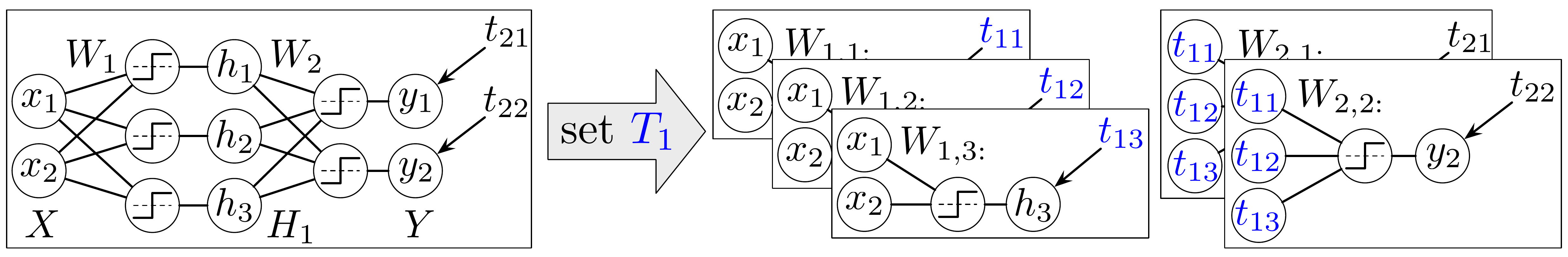}
\caption{\small{
After setting the hidden-layer targets $T_1$ of a deep hard-threshold network, 
the network decomposes into independent perceptrons, which can then be learned with
standard methods. 
\vspace{-1.5mm}
}}
\label{fig:mlp_decomp}
\end{centering}
\end{figure}

Consider a simple single-hidden-layer hard-threshold network 
$Y = f(X ; W) = g(W_2 ~g(W_1 X)) = g(W_1 H_1)$ for a dataset $\mc{D} = (X, T_2)$,
where $H_1 = g(W_1 X)$ are the hidden-layer activations.
An example of such a network is shown on the left side of Figure~\ref{fig:mlp_decomp}.
Clearly, $Y$ and $H_1$ are both collections of (single-layer) perceptrons.
Backpropagation cannot be used to train the input layer's
weights $W_1$ because of the hard-threshold activations 
but, since each hidden activation $h_{1j}$
is the output of a perceptron,
if we knew the value $t_{1j} \in \{-1, +1\}$
that each hidden unit \textit{should} take for each input $\x^\ds{i}$,
we could then use the perceptron algorithm to set the 
first-layer weights, $W_1$, to produce these target values.
We refer to $t_{1j}$ as the \textit{target} of $h_{1j}$.
Given a matrix of hidden-layer targets 
$T_1 \in \{-1, +1\}^{n_1 \times m}$, each layer 
(and in fact each perceptron in each layer) can be
learned separately, as they no longer depend on each other,
where the goal of perceptron learning is to update the weights of each layer $d$
so that its activations $H_d$ equal its targets $T_d$ given inputs $T_{d-1}$.
Figure~\ref{fig:mlp_decomp} shows an example of this decomposition.
We denote the targets of an $\ell$-layer network as $T = \{ T_1, \dots, T_\ell \}$,
where $T_k$ for $k = 1 \dots \ell-1$ are the hidden-layer targets and
$T_\ell$ are the dataset targets. 
We often let $T_0 = X$ for notational convenience.

Auxiliary-variable-based approaches, such as ADMM~\citep{Taylor2016a,Carreira-Perpinan2012a}
and other target propagation methods~\citep{LeCun1986,Lee2015}
use a similar process for decomposing the layers of a network; however, these
focus on continuous variables and impose (soft) constraints
to ensure that each activation equals its auxiliary variable.
We take a different approach here, inspired by the combinatorial nature of the 
problem and the perceptron algorithm.

Since the final layer is a perceptron, 
the training instances can only be separated
if the hidden-layer activations $H_1$
are linearly separable with respect to the dataset targets $T_2$.
Thus, the hidden-layer targets $T_1$ must be set such that they are linearly
separable with respect to the dataset targets $T_2$, since the hidden-layer targets $T_1$
are the intended values of their activations $H_1$.
However, in order to ensure
that the hidden-layer activations $H_1$ will equal their targets $T_1$ after
training,
the hidden-layer targets $T_1$ must be able to be produced (exactly) by the first layer,
which is only possible if the hidden-layer targets $T_1$ are also linearly separable
with respect to the inputs $X$. 
Thus, a sufficient condition for $f(X;W)$ to separate the data is that the
hidden-layer targets induce linear separability in all units in both layers of the network.
We refer to this property as \textit{feasibility.} \looseness=-1

\begin{define}
A setting of the targets $T = \{T_1, \dots, T_\ell\}$ of an $\ell$-layer 
deep hard-threshold network
$f(X ; W)$ is \emph{feasible} for a dataset $\mc{D} = (X, T_\ell)$ iff 
for each unit $j = 1 \dots n_d$ in each layer $d = 1 \dots \ell$
the dataset formed by its inputs $T_{d-1}$ and targets $T_{d,j:}$ 
is linearly separable, where $T_0 \triangleq X$.
\end{define}

Feasibility is a much weaker condition than linear separability, since
the output decision boundary of a multilayer hard-threshold network 
with feasible targets is in general highly nonlinear.
It follows from the definition of feasibility and 
convergence of the perceptron algorithm
that if a feasible setting of a network's targets on a dataset
exists, the network can separate the training data.

\begin{prop}
Let $\mc{D} = \{(\x^\ds{i}, t^\ds{i})\}$ be a dataset and let 
$f(X ; W)$ be an $\ell$-layer hard-threshold network with feasible targets 
$T = \{ T_1, \dots, T_\ell \}$ in which each layer $d$ of $f$ was trained separately
with inputs $T_{d-1}$ and targets $T_d$, where $T_0 \triangleq X$, 
then $f$ will correctly classify
each instance $\x^\ds{i}$, such that 
$f(\x^\ds{i} ; W) t^\ds{i} > 0$ for all $i = 1 \dots m$.
\end{prop}


Learning a deep hard-threshold network thus reduces to 
finding a feasible setting of its targets
and then optimizing its weights given these targets, 
i.e., mixed convex-combinatorial optimization.
The simplest method for this 
is to perform exhaustive search on the targets. Exhaustive search 
iterates through all possible settings of the hidden-layer targets, 
updating the weights of each perceptron
whose inputs or targets changed, and returns the weights
and feasible targets that result in the lowest loss.
While impractical, exhaustive search is worth briefly
examining to better understand the solution space.
In particular, because of the decomposition afforded by setting the targets,
exhaustive search over just the targets is sufficient
to learn the globally optimal deep hard-threshold network, even though the weights
are learned by gradient descent.

\begin{prop}
If a feasible setting of a deep hard-threshold network's targets on a 
dataset $\mc{D}$ exists, 
then exhaustive search 
returns the global minimum of the loss in time exponential 
in the number of hidden units.
\end{prop}

Learning can be improved and feasibility 
relaxed if, instead of the perceptron algorithm, 
a more robust method is used for perceptron learning.
For example, a perceptron can be learned for a 
non-linearly-separable dataset
by minimizing the hinge loss $L(z,t) = \max(0, 1-tz)$,
a convex loss on the perceptron's pre-activation output $z$ and target $t$
that maximizes the margin when combined with L2 regularization.
In general, however, 
any method for learning linear classifiers can be used.
We denote the loss used to train the weights of a layer $d$
as $L_d$, where the loss of the final layer $L_\ell$ is
the output loss. \looseness=-1

At the other end of the search spectrum is hill climbing.
In each iteration, hill climbing evaluates all neighboring states 
of the current state (i.e., target settings that differ
from the current one by only one target) 
and chooses the one with the lowest loss. 
The search halts when none of the new states improve the loss.
Each state is evaluated by 
optimizing the weights of each perceptron given the state's targets,
and then computing the output loss.
Hill climbing is more practical than exhaustive search,
since it need not explore an exponential number of states, and it
also provides the same local optima guarantee as 
gradient descent 
on soft-threshold networks.
\looseness=-1

\begin{prop}
Hill climbing on the targets of a deep hard-threshold network
returns
a local minimum of the loss, where each iteration takes time
linear in the size of the set of proposed targets. 
\end{prop}


Exhaustive search and hill climbing comprise two ends of the discrete
optimization spectrum. 
Beam search, which 
maintains a beam of the most promising solutions and explores each,
is another powerful approach that
contains both hill climbing and exhaustive search as special cases.
In general, however, any discrete
optimization algorithm can be used for setting targets. 
For example, methods from satisfiability solving, integer linear programming, or 
constraint satisfaction might work well,
as the linear separability requirements of feasibility
can be viewed as constraints on the search space.


We believe that our mixed convex-combinatorial optimization 
framework opens many 
new avenues for developing learning algorithms for deep networks,
including those with non-differentiable modules.
In the following section, we use these ideas to develop a 
learning algorithm that hews much closer to standard methods, 
and in fact contains the straight-through estimator as a special case.

\vspace{-1mm}
\section{Feasible target propagation} 
\label{sec:ftprop}
\vspace{-1mm}

The open question from the preceding section
is how to set the hidden-layer targets.
Generating good, feasible targets for the entire network
at once is a difficult problem; instead, an easier approach is
to propose targets for only one layer at a time. 
As in backpropagation, it makes sense to start from the output layer,
since the final-layer targets are given,
and successively set targets for each upstream layer.
Further, since it is hard to know a priori if a setting
of a layer's targets is feasible for a given network architecture,
a simple alternative is to set the targets for a layer $d$
and then optimize the upstream weights 
(i.e., weights in layers $j \leq d$ ) to check if the targets are feasible. 
Since the goals when optimizing a layer's weights and when setting its upstream targets
(i.e., its inputs) are the same -- namely, to induce feasibility --
a natural method for setting target values is to choose targets
that reduce the layer's loss $L_d$. 
However, because the targets are discrete, moves in target space
are large and non-smooth and cannot be guaranteed to lower the loss
without actually performing the move.
Thus, heuristics are necessary.
We discuss these in more detail below.

Determining feasibility of the targets at layer $d$ can be done by recursively
updating the weights of layer $d$ and
proposing targets for layer $d-1$ given the targets for layer $d$.
This recursion continues until the input layer is reached, where
feasibility (i.e., linear separability) can be easily determined
by optimizing that layer's weights given its targets and the dataset inputs. 
The targets at layer $d$ can then be
updated based on the information gained from the recursion
and, if the upstream weights were altered, based on the new outputs of layer $d-1$.
We call this recursive algorithm \emph{feasible target propagation}, or \algname{}.
Pseudocode is shown in Algorithm~\ref{alg:ftprop}. 

\newcommand*\Let[2]{#1 $\gets$ #2}
\algrenewcommand\algorithmicrequire{\textbf{Input:}}
\algrenewcommand\algorithmicensure{\textbf{Output:}}
\renewcommand{\algorithmiccomment}[1]{\bgroup\hfill\footnotesize //~\emph{{#1}}\egroup}

\algnewcommand{\IIf}[1]{\State\algorithmicif\ #1\ \algorithmicthen}
\algnewcommand{\ElseIIf}[1]{\State\algorithmicelsif\ #1\ \algorithmicthen}
\algnewcommand{\ElseI}[1]{\State\algorithmicelse\ #1\ }
\algnewcommand{\EndIIf}{\unskip\ }

\algrenewcommand\algorithmicindent{1.2em}%
\makeatletter
\newcommand{\mybox}[2][\fboxsep]{{%
  \setlength{\fboxsep}{#1}\fbox{\m@th$\displaystyle#2$}}}
\makeatother

\begin{algorithm*}[htb]
  \caption{Train an $\ell$-layer hard-threshold network $Y = f(X ; W)$ 
           on dataset ${\mc{D} = (X, T_\ell)}$ with feasible target propagation (\algname{})
           using loss functions \hspace{0.2em}$L = \{L_d\}_{d=1}^\ell$.}
  \label{alg:ftprop} 
  \begin{algorithmic}[1]
    \State initialize weights $W = \{W_1, \dots, W_\ell\}$ randomly
    \State initialize targets $T_1, \dots, T_{\ell-1}$ as the outputs of their hidden units in $f(X ; W)$ 
    \State set $T_0 \gets X$ and set $T \gets \{T_0, T_1, \dots, T_\ell \}$
    \State \algname{}$(W, T, L, \ell)$  \Comment{train the network by searching for a feasible target setting}

    \vspace{0.5em}

    \Function{\algname{}}{weights $W$, targets $T$, losses $L$, and layer index $d$}
    \State optimize $W_d$ with respect to layer loss $L_d(Z_d, T_d)$ \Comment{check feasibility; $Z_d = W_d T_{d-1}$}
    \If{activations $H_d = g(W_d T_{d-1})$ equal the targets $T_d$} \Return True \Comment{feasible}
    \ElsIf{this is the first layer (i.e., $d=1$)} \Return False \Comment{infeasible}
    \EndIf
    \While{computational budget of this layer not exceeded} \Comment{e.g., determined by beam search}
      \State $T_{d-1} \gets $ heuristically set targets for upstream layer to reduce layer loss $L_d(Z_d, T_{d})$ 
      \If{\algname{}$(W, T, L, d-1)$} \Comment{check if targets $T_{d-1}$ are feasible}
        \State optimize $W_d$ with respect to layer loss $L_d(Z_d, T_d)$  
        \If{activations $H_d = g(W_d T_{d-1})$ equal the targets $T_d$} \Return True \Comment{feasible}
        \EndIf
      \EndIf
    \EndWhile
    \State \Return False
    \EndFunction
  \end{algorithmic}
\end{algorithm*}

As the name implies, \algname{} is a form of 
target propagation~\citep{LeCun1986,LeCun1987,Lee2015}
that uses discrete optimization to set discrete targets, instead of 
using continuous optimization to set continuous targets.
\algname{} is also highly related to RDIS~\citep{Friesen2015},
a powerful nonconvex optimization algorithm based on 
satisfiability (SAT) solvers that recursively chooses and sets subsets of variables in order
to decompose the underlying problem into simpler subproblems.
While RDIS is applied only to continuous problems, 
the ideas behind RDIS can be generalized to discrete variables
via the sum-product theorem~\citep{Friesen2016}. 
This suggests an interesting connection between \algname{}
and SAT that we leave for future work.

%
%
%

Of course, modern deep networks will not always have a feasible
setting of their targets for a given dataset. 
For example,
a convolutional layer imposes a large amount of structure on its weight matrix, 
making it less likely that the layer's input will be linearly separable with respect
to its targets. 
Further, ensuring feasibility will in general cause learning to overfit the training data, 
which will worsen generalization performance.
Thus, we would like to relax the feasibility requirements.

In addition, there are many 
benefits of using mini-batch instead of full-batch training,
including improved generalization gap (e.g., see \citet{LeCun2012} or \citet{Keskar2016}), 
reduced memory usage, the ability to exploit data augmentation, and the
prevalence of tools 
(e.g., GPUs) designed for it. 
\looseness=-1

Fortunately, it is straightforward to convert \algname{} to a mini-batch algorithm
and to relax the feasibility requirements. 
In particular, since it is important not to overcommit to any one
mini-batch, the mini-batch version of \algname{} 
(i) only updates the weights and targets of each layer once per mini-batch;
(ii) only takes a small gradient step on each layer's weights, instead of optimizing them fully;
(iii) sets the targets of the downstream layer in parallel with updating the current layer's weights,
since the weights will not change much; and (iv) removes all checks for feasibility.
We call this algorithm \algnamemb{} and present pseudocode in Algorithm~\ref{alg:ftpropmb}.
\algnamemb{} closely resembles backpropagation-based
methods, 
allowing us to easily implement it with
standard libraries.
\looseness=-1

\begin{algorithm*}[thb]
  \caption{Train an $\ell$-layer hard-threshold network $Y = f(X ; W)$ 
          on dataset ${\mc{D} = (X, T_\ell)}$ with mini-batch feasible target propagation (\algnamemb)
          using loss functions $L = \{L_d\}_{d=1}^\ell$.}
  \label{alg:ftpropmb} 
  \begin{algorithmic}[1]
        \State initialize weights $W = \{W_1, \dots, W_\ell\}$ randomly
        \For{each minibatch $(X_b, T_b)$ from $\mc{D}$}
          \State initialize targets $T_1, \dots, T_{\ell-1}$ as the outputs of their hidden units in $f(X_b ; W)$ \Comment{forward pass}
          \State set $T_0 \gets X_b$, set $T_\ell \gets T_b$, and set $T \gets \{ T_0, \dots, T_\ell\}$
          \State \algnamemb{}$(W, T, L, \ell)$
        \EndFor
        \vspace{0.5em}
 
  \Function{\algnamemb{}}{weights $W$, targets $T$, losses $L$, and layer index $d$}
      \State $\hat{T}_{d-1} \gets$ set targets for upstream layer based on current weights $W_d$ and loss $L_d(Z_d, T_d)$
      \State update $W_d$ with respect to layer loss $L_d(Z_d, T_d)$ \Comment{where $Z_d = W_d T_{d-1} = W_d H_{d-1}$}
      \If{$d > 1$} \algnamemb{}$(W, ~\{T_0, \dots, \hat{T}_{d-1}, \dots, T_\ell \}, ~L, ~d-1)$
      \EndIf
    \EndFunction
    
  \end{algorithmic}
\end{algorithm*}

\vspace{-1mm}
\subsection{Target heuristics}
\vspace{-1mm}

\newcommand{\heur}[1]{{r(#1)}}

When the activations of each layer are differentiable, backpropagation
provides a method for telling each layer how to adjust its 
outputs to improve the loss. 
Conversely, in hard-threshold networks, target propagation provides
a method for telling each layer how to adjust its 
outputs to improve the next layer's loss.
While gradients cannot propagate through hard-threshold units, the
derivatives within a layer can still be computed.
An effective and efficient heuristic for setting the target $t_{dj}$ for an activation $h_{dj}$ of layer $d$ is
to use the (negative) sign of the partial derivative of the next layer's loss. 
Specifically, we set $t_{dj} = \heur{h_{dj}}$, where
\begin{align}
\heur{h_{dj}} \triangleq \sign{\left(- \frac{\partial }{\partial h_{dj}}L_{d+1}(Z_{d+1}, T_{d+1})\right)}
\label{eqn:heur}
\end{align}
and $Z_{d+1}$ is either the pre-activation or post-activation output, 
depending on the choice of loss.

When used to update only a single target at a time, this heuristic will often 
set the target value that correctly results in the lowest loss.
In particular, when $L_{d+1}$ is convex, its negative partial derivative with respect to $h_{dj}$ 
by definition points in the direction of the global minimum of $L_{d+1}$. 
Without loss of generality, let $h_{dj} = -1$.
Now, if 
$\heur{h_{dj}}  = -1$, then it follows from the convexity of the loss 
that flipping $h_{dj}$ and keeping all other variables the same would increase $L_{d+1}$.
On the other hand, if $\heur{h_{dj}} = +1$, 
then flipping $h_{dj}$ may or may not reduce the loss, since convexity cannot tell us
which of $h_{dj} = +1$ or $h_{dj} = -1$ results in a smaller $L_{d+1}$.
However, the discrepancy between $h_{dj}$ and $\heur{h_{dj}}$ 
indicates a lack of confidence in the current value of $h_{dj}$.
A natural choice is thus to set $t_{dj}$ to push the pre-activation
value of $h_{dj}$ towards $0$, making $h_{dj}$ more likely to flip.
Setting $t_{dj} = \heur{h_{dj}} = +1$ accomplishes this.
We note that, while this heuristic performs well, 
there is still room for improvement, for example
by extending $\heur{\cdot}$ to better handle the $h_{dj} \neq \heur{h_{dj}}$ case
or by combining information across the batch. We leave such investigations for future work.



\vspace{-1mm}
\subsection{Layer loss functions}
\vspace{-1mm}

\begin{figure}[t]
  \centering
  \begin{subfigure}[b]{0.24\columnwidth}
    \includegraphics[width=\textwidth]{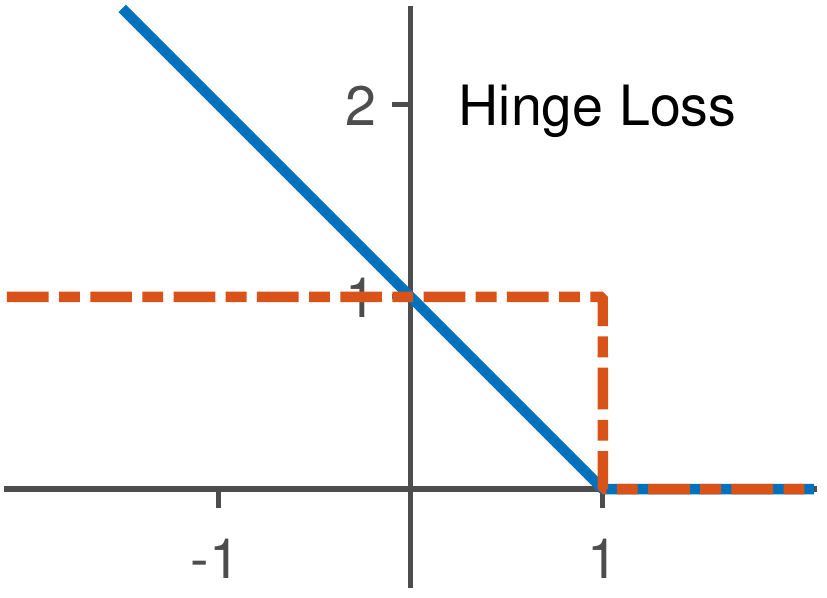}
    \caption{}
    \label{fig:hinge}
  \end{subfigure}
  \hspace{0.05em}
  \begin{subfigure}[b]{0.24\columnwidth}
    \includegraphics[width=\textwidth]{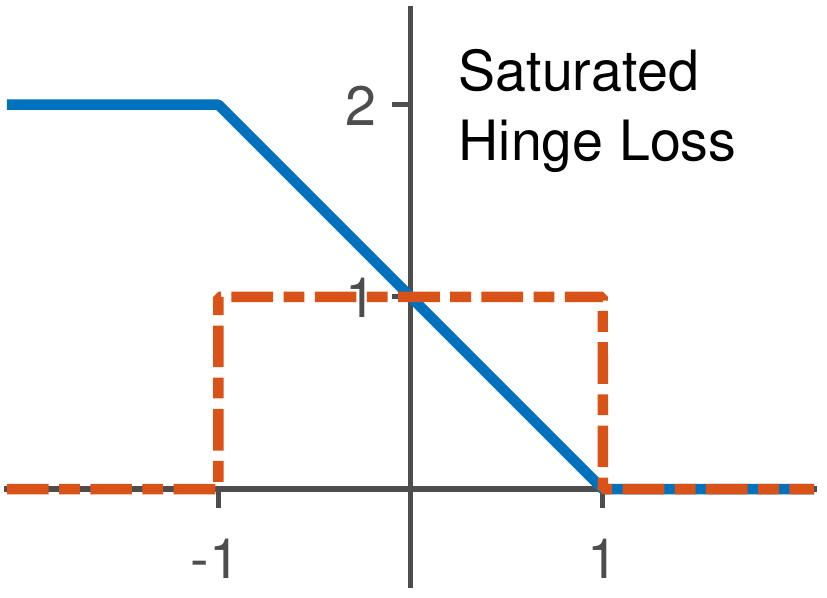}
    \caption{}
    \label{fig:sat_hinge}
  \end{subfigure}
  \hspace{0.05em}
  \begin{subfigure}[b]{0.24\columnwidth}
    \includegraphics[width=\textwidth]{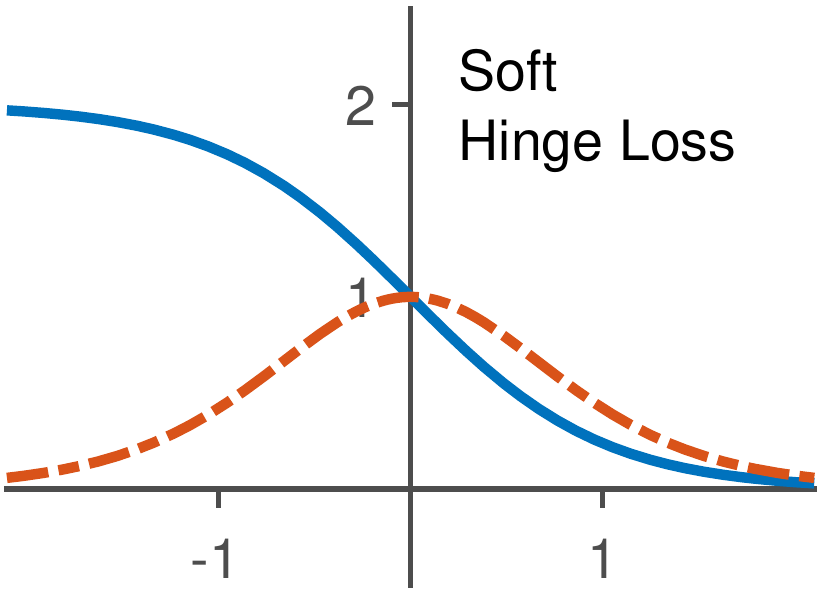}
    \caption{}
    \label{fig:soft_hinge}
  \end{subfigure}
  \hspace{0.05em}
  \begin{subfigure}[b]{0.24\columnwidth}
    \includegraphics[width=\textwidth]{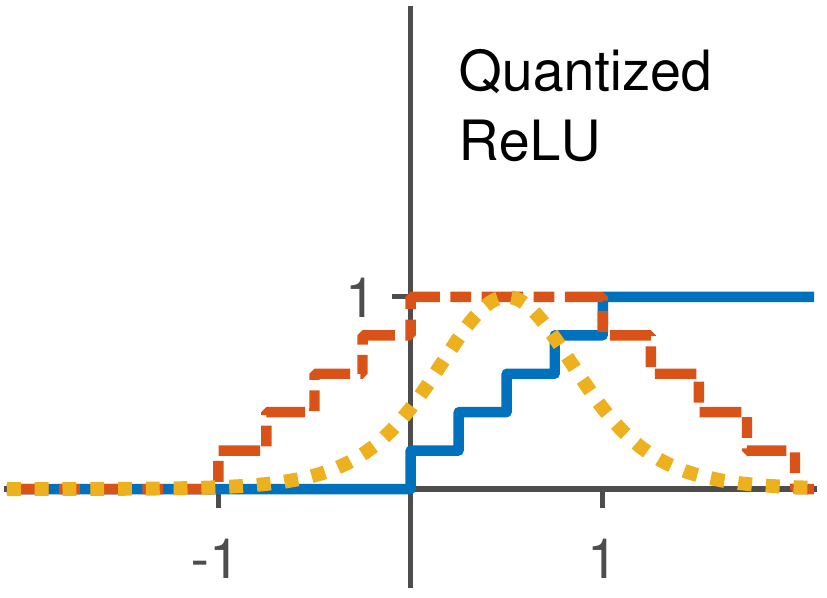}
    \caption{}
    \label{fig:staircase}
  \end{subfigure}
  \caption{\small{
                Figures (a)-(c) show different per-layer loss functions (solid blue line) 
                and their derivatives (dashed red line). 
                Figure (d) shows the quantized ReLU activation (solid blue line), 
                which is a sum of step functions, 
                its corresponding sum of saturated-hinge-loss derivatives (dashed red line),
                and the soft-hinge-loss approximation to this sum that was found to work best (dotted yellow line).
        }
		}
\end{figure}

The hinge loss, shown in Figure~\ref{fig:hinge}, is a robust version of the
perceptron criterion and is thus a natural per-layer loss function to use
for finding good settings of the targets and weights, even when there are no
feasible target settings. 
However, in preliminary experiments we found that learning tended to stall
and become erratic over time when using the hinge loss for each layer.
We attribute this to two separate issues.
First, the hinge loss is sensitive to noisy data and outliers~\citep{Wu2007},
which can cause learning to focus on instances that are unlikely to ever
be classified correctly, instead of on instances near the separator.
Second, since with convolutional layers and large, noisy datasets 
it is unlikely that a layer's inputs are entirely linearly separable, 
it is important to prioritize some targets over others. 
Ideally, the highest priority targets would be those with the largest
effect on the output loss.
\looseness=-1

The first issue can be solved by saturating (truncating) the hinge loss,
thus making it less sensitive to outliers~\citep{Wu2007}.
The saturated hinge loss, shown in Figure~\ref{fig:sat_hinge},
is $\sathinge(z, t; b) = \max(0, 1- \max(tz, b))$
for some threshold $b$, where we set $b=-1$ to make its derivative symmetric.
The second 
problem can be solved in a variety of ways, including randomly
subsampling targets or weighting the loss associated with each target
according to some heuristic.
The simplest and most accurate method that we have found is to
weight the loss for each target $t_{dj}$ by the magnitude of the partial derivative
of the next layer's loss $L_{d+1}$ with respect to the target's hidden unit $h_{dj}$,
such that 
\begin{align}
L_d(z_{dj}, t_{dj}) = \sathinge(z_{dj}, t_{dj}) \cdot \left| 
\frac{\partial L_{d+1}}{\partial h_{dj}} 
\right|.
\label{eqn:lloss}
\end{align}

While the saturated hinge loss works well, 
if the input $z_{dj}$ ever moves out of the range $[-1, +1]$
then its derivative will become zero and the unit will no longer be trainable. 
To avoid this, we propose the \emph{soft hinge loss}, 
shown in Figure~\ref{fig:soft_hinge}, where $\softhinge(z, t) = \tanh(-tz) + 1$.
Like the saturated hinge, the soft hinge has slope $1$ at the threshold and
has a symmetric derivative; however, it also benefits from having a larger
input region with non-zero derivative.
Note that \citet{Bengio2013} report that using the derivative of a sigmoid
as the STE performed worse than the identity function.
Based on our experiments with other loss functions, 
including variations of the squared hinge loss
and the log loss, this is most likely because the slope of the sigmoid
is less than unity at the threshold, which causes vanishing gradients.
Loss functions with asymmetric derivatives around the threshold
also seemed to perform worse than those with symmetric derivatives
(e.g., the saturating and soft hinge losses).
In our experiments, we show that the soft hinge loss outperforms
the saturated hinge loss for both sign  
and quantized ReLU activations, which we discuss below.

\vspace{-1mm}
\subsection{Relationship to the straight-through estimator}
\vspace{-1mm}

When each loss term in each hidden layer is scaled by the magnitude of the
partial derivative of its downstream layer's loss and each target 
is set based on the sign of the same partial derivative,
then target propagation transmits information about the output loss to every layer
in the network, despite the hard-threshold units.
Interestingly, this combination of loss function and target heuristic
can exactly reproduce the weight updates of the straight-through estimator (STE).
Specifically, 
the weight updates that result from using 
the scaled saturated hinge loss from~(\ref{eqn:lloss}) and
the target heuristic in~(\ref{eqn:heur}) are exactly those of the
saturated straight-through
estimator (SSTE) defined in \citet{Hubara2016a}, which replaces the derivative
of $\sign(z)$ with $1_{|z| \leq 1}$, where $1_{(\cdot)}$ is the indicator function.
Other STEs correspond to different choices of per-layer loss function.
For example, the original STE corresponds to the linear loss $L(z,t) = -tz$
with the above target heuristic.
This connection provides a justification for existing STE approaches, which can now
each be seen as an instance of \algname{} 
with a particular choice of per-layer loss function and target heuristic. 
We believe that this will enable more principled investigations
and extensions of these methods in future work.

%
\vspace{-1mm}
\subsection{Quantized activations}
\vspace{-1mm}

Straight-through estimation is also commonly used
to backpropagate through quantized variants of standard activations, such as the ReLU.
Figure~\ref{fig:staircase} shows a quantized ReLU (qReLU) with $6$ evenly-spaced quantization levels.
The simplest and most popular straight-through estimator (STE) for qReLU 
is to use the derivative of the
saturated (or clipped) ReLU $\frac{\partial \satrelu(x)}{\partial x} = 1_{0 < x < 1}$, 
where $\satrelu(x) = \min(1, \max(x, 0))$.
However, if we instead consider the qReLU activation from the viewpoint of 
\algname{}, 
then the qReLU becomes a (normalized) 
sum of step functions $\qrelu(z) = \frac{1}{k} \sum_{i=0}^{k-1} \step(z - \frac{i}{k-1})$, 
where $\step(z) = 1$ if $z > 0$ and $0$ otherwise, and is a linear transformation of $\sign(z)$.
The resulting derivative of the sum of saturated hinge losses (one for each step function) is shown
in red in Figure~\ref{fig:staircase}, and is clearly quite different than the STE described above.
In initial experiments, this performed as well as or better than the STE;
however, we achieved additional performance improvements by using the 
softened approximation shown in yellow in Figure~\ref{fig:staircase}, which is simply
the derivative of a soft hinge that has been scaled and shifted to match the
qReLU domain. This is a natural choice because the derivative of a sum of a small number of soft hinge losses
has a shape similar to that of the derivative of a single soft hinge loss.

\vspace{-1mm}
\section{Experiments}
\label{sec:exp}
\vspace{-1mm}

We evaluated \algnamemb{} with soft hinge per-layer losses (FTP-SH) for
training deep networks with sign and 2- and 3-bit qReLU activations by comparing models
trained with FTP-SH to those trained with the saturated straight-through 
estimators (SSTEs) described earlier
(although, as discussed, these SSTEs can also be seen as instances of \algnamemb{}).
We compared to these SSTEs because they are the standard approach
in the literature and they significantly outperformed the STE
in our initial experiments 
(\citet{Hubara2016a} observed similar behavior).
Computationally, \algnamemb{} has the same performance as 
straight-through estimation; however, the soft hinge
loss involves computing a hyperbolic tangent, which requires more computation than
a piecewise linear function.
This is the same performance difference seen 
when using sigmoid activations instead of ReLUs in soft-threshold networks.
We also trained each model with ReLU and 
saturated-ReLU activations as full-precision baselines.
\looseness=-1

We did not use weight quantization because our main interest is
training with hard-threshold activations, 
and because recent work has shown that
weights can be quantized with little effect on 
performance~\citep{Hubara2016a,Rastegari2016,Zhou2016}.
We tested these training methods on the 
CIFAR-10~\citep{Krizhevsky09} and ImageNet (ILSVRC 2012)~\citep{ILSVRC15} datasets. 
On CIFAR-10, we trained a simple $4$-layer convolutional network
and the $8$-layer convolutional network of~\citet{Zhou2016}.
On ImageNet, we trained AlexNet~\citep{Krizhevsky2012}, 
the most common model in the quantization literature, 
and ResNet-18~\citep{He2015}.
Further experiment details 
are provided in Appendix~\ref{apx:exp}, 
along with learning curves for all experiments.
Code is available at {\small \url{https://github.com/afriesen/ftprop}}.

%


\vspace{-1mm}
\subsection{CIFAR-10}
\vspace{-1mm}

%

\newcommand{\pct}{}
\newcommand{\rulecolor}[1]{%
  \def\CT@arc@{\color{#1}}%
}
\definecolor{mygray}{gray}{0.25}

\begin{table}[tb]
\centering
\caption{\small{The best top-1 test accuracy 
for each network over all epochs when trained with sign, qReLU, and full-precision 
baseline activations on CIFAR-10 and ImageNet. 
The hard-threshold activations are trained with both \algnamemb{} with 
per-layer soft hinge losses (FTP-SH) and the 
saturated straight-through estimator (SSTE).
Bold numbers denote the best performing quantized activation in each experiment.
\vspace{-0.35em}
}}
\begin{tabular}{@{} l *7c *2c @{}}
\toprule
\multicolumn{1}{l}{\multirow{2}{*}{}} & \hspace{0em} & 
\multicolumn{2}{c}{{Sign}} & \hspace{0em} & 
\multicolumn{2}{c}{{qReLU}}  & \hspace{0em} &
\multicolumn{2}{c}{\color{mygray} Baselines} \\
 \cmidrule[.5\cmidrulewidth](lr){3-4} \cmidrule[.5\cmidrulewidth](lr){6-7}
 \cmidrule[.5\cmidrulewidth](l){9-10}
 & & \emph{SSTE} & \emph{FTP-SH} & & \emph{SSTE} & \emph{FTP-SH} & & \emph{\color{mygray} ReLU} & \emph{\color{mygray} Sat. ReLU} \\ 
\cmidrule[2\cmidrulewidth](r){1-1} \cmidrule[2\cmidrulewidth](lr){3-4} \cmidrule[2\cmidrulewidth](lr){6-7} \cmidrule[2\cmidrulewidth](l){9-10}
{4-layer convnet (CIFAR-10)} && 80.6\pct & \textbf{81.3}\pct & & 85.6\pct & 85.5\pct & & {\color{mygray}86.5\pct} & {\color{mygray}87.3\pct}  \\
\cmidrule(r){1-1} \cmidrule(lr){3-4} \cmidrule(lr){6-7} \cmidrule(l){9-10}
{8-layer convnet (CIFAR-10)} & & 84.6\pct & \textbf{84.9}\pct & & 88.4\pct & \textbf{89.8}\pct & & {\color{mygray} 91.2\pct} & {\color{mygray} 91.2\pct}  \\
\cmidrule[3\cmidrulewidth](r){1-1} \cmidrule[3\cmidrulewidth](lr){3-4} \cmidrule[3
\cmidrulewidth](lr){6-7} \cmidrule[3\cmidrulewidth](l){9-10}
{AlexNet (ImageNet)} & & 46.7\pct & \textbf{47.3}\pct & & 59.4\pct & \textbf{60.7}\pct & & {\color{mygray} 61.3\pct} & {\color{mygray}61.9\pct}  \\
\cmidrule(r){1-1} \cmidrule(lr){3-4} \cmidrule(lr){6-7} \cmidrule(l){9-10}
{ResNet-18 (ImageNet)} & & \textbf{49.1}\pct & 47.8\pct & & 60.6\pct & \textbf{64.3}\pct & & {\color{mygray}69.1\pct} & {\color{mygray}66.9\pct}  \\
\bottomrule
\end{tabular}
\label{table:accuracies}
\end{table}

Test accuracies for the 4-layer and 8-layer convolutional networks on 
CIFAR-10 are shown in Table~\ref{table:accuracies}.
For the 4-layer model, FTP-SH shows a consistent $0.5$-$1$\% 
accuracy gain over SSTE 
for the entire training trajectory, resulting in the $0.7$\% 
improvement shown in Table~\ref{table:accuracies}.
However, for the 2-bit qRELU activation, SSTE and FTP-SH perform 
nearly identically in the 4-layer model.
Conversely, for the more complex 8-layer model, the FTP-SH accuracy 
is only $0.3$\% above SSTE for the sign activation, but
for the qReLU activation FTP-SH achieves a consistent $1.4$\% 
improvement over SSTE.

We posit that the decrease in performance gap for the sign activation when moving 
from the 4- to 8-layer model is because both methods are able to 
effectively train the higher-capacity
model to achieve close to its best possible performance on this 
dataset, whereas the opposite is true
for the qReLU activation; i.e., the restricted capacity of the 
4-layer model limits the ability of both
methods to train the more expressive qReLU effectively. 
If this is true, 
then we expect that FTP-SH will outperform SSTE for both the sign
and qReLU activations on a harder dataset.
Unsurprisingly, none of the low-precision methods perform 
as well as the baseline high-precision methods; 
however, the narrowness of the performance gap between 2-bit qReLU with FTP-SH and 
full-precision ReLU is encouraging.
\looseness=-1

\vspace{-1mm}
\subsection{ImageNet}
\vspace{-1mm}

\begin{figure}[tb]
  \centering
  \begin{subfigure}[b]{0.49\columnwidth}
    \includegraphics[width=\textwidth]{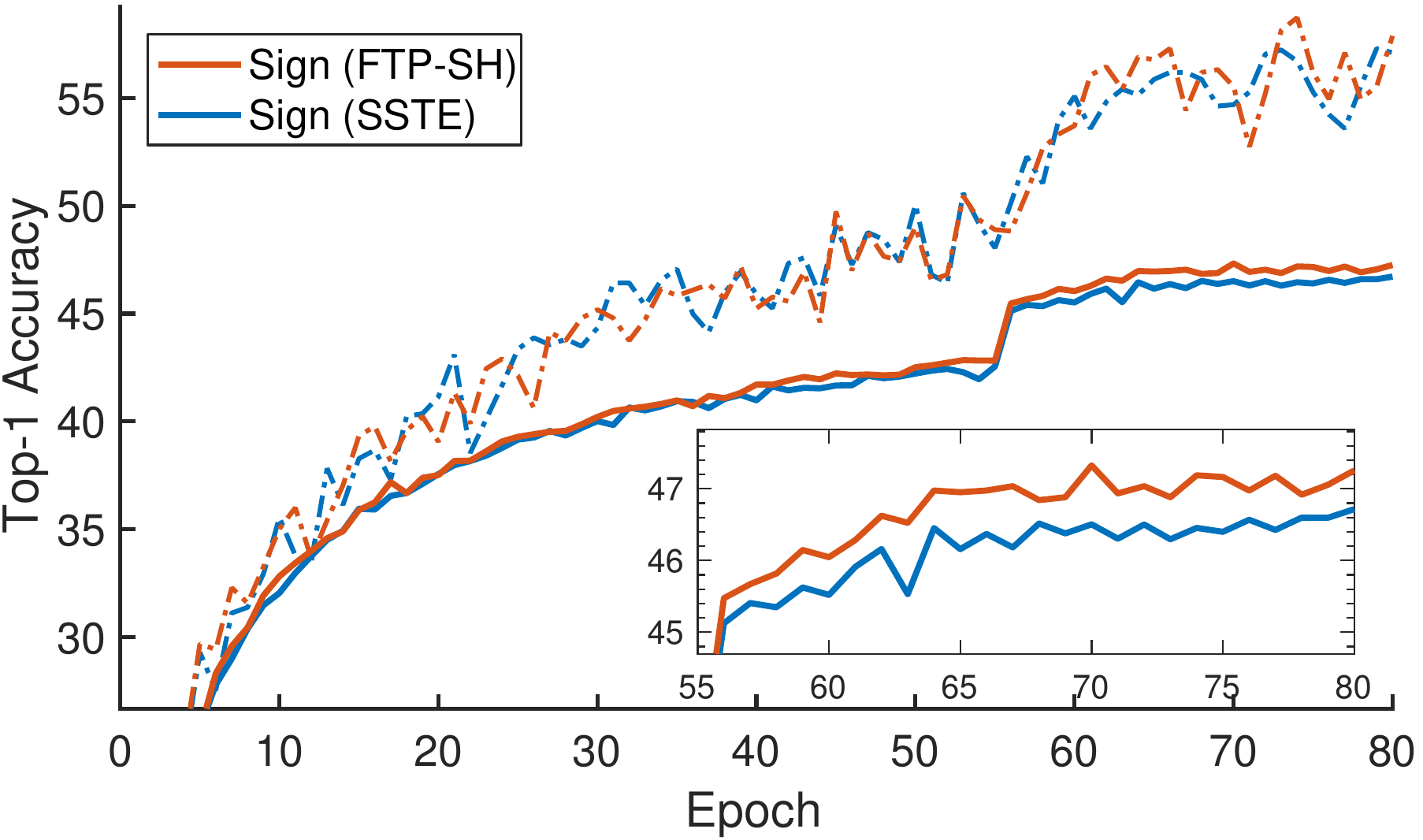}
  \end{subfigure}
  \begin{subfigure}[b]{0.49\columnwidth}
    \includegraphics[width=\textwidth]{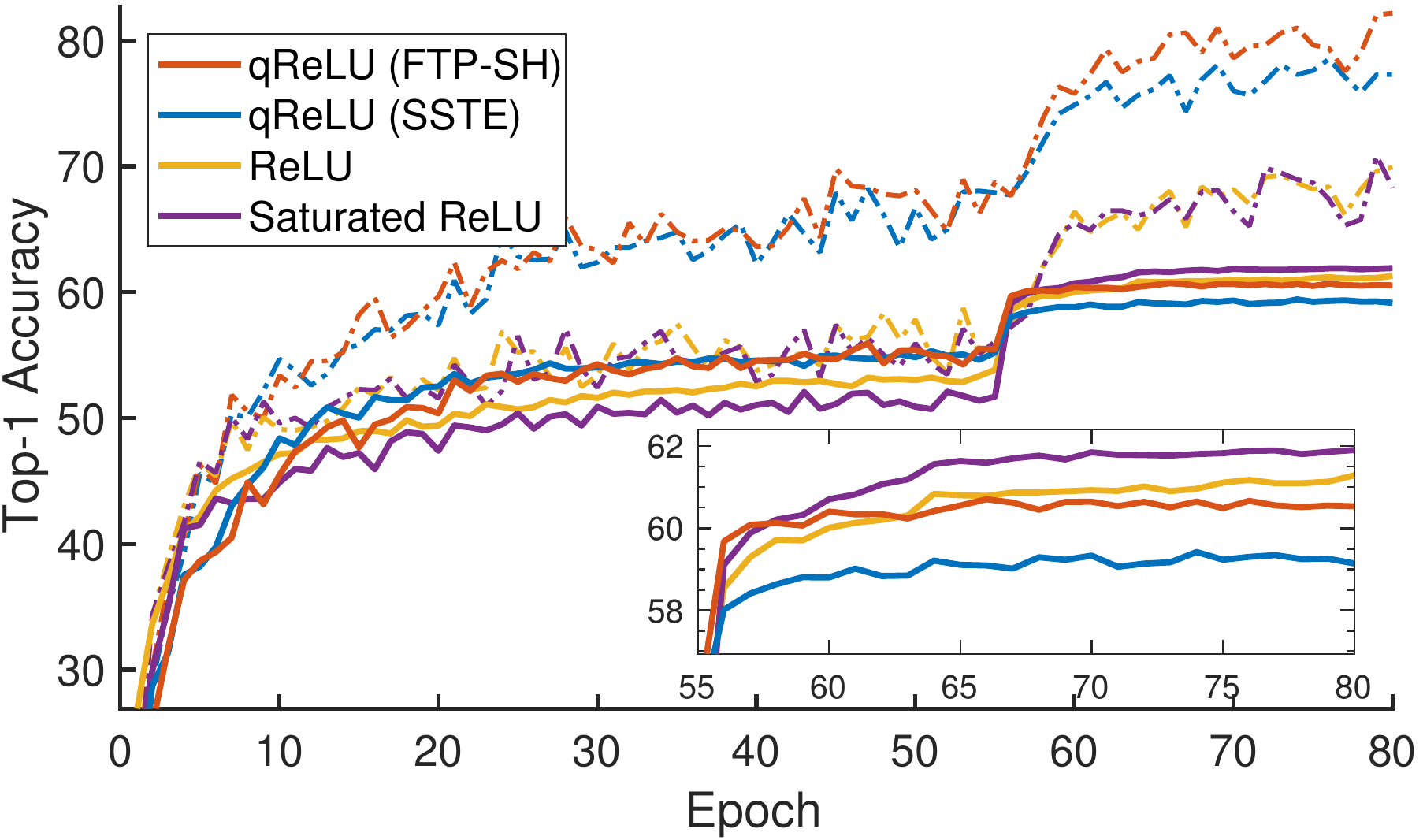}
  \end{subfigure}
  \hspace{0.05em}
  \caption{	
  \small{
%
The top-1 train (thin dashed lines) and test (thicker solid lines) accuracies for AlexNet with
different activation functions on ImageNet. 
The inset figures show the test accuracy for the final $25$ epochs in detail. 
In both figures, \algnamemb{} with soft hinge (FTP-SH, red) outperforms the 
saturated straight-through estimator (SSTE, blue). 
The left figure shows the network with sign activations.
The right figure shows that the 2-bit quantized ReLU (qReLU) trained with our method (FTP-SH) performs
nearly as well as the full-precision ReLU. Interestingly, saturated ReLU outperforms standard ReLU.
Best viewed in color.
\vspace{-0.15em}
}}
\label{fig:imagenet} 
\end{figure}

%
The results from the ImageNet experiments are also shown in Table~\ref{table:accuracies}.
As predicted from the CIFAR-10 experiments, we see that FTP-SH improves 
test accuracy on AlexNet for both sign and 2-bit qReLU activations on the more challenging
ImageNet dataset. 
This is also shown in Figure~\ref{fig:imagenet}, which plots the top-1
train and test accuracy curves for the six different activation functions 
for AlexNet on ImageNet.
The left-hand plot shows that training sign activations with FTP-SH provides consistently
better test accuracy than SSTE throughout the training trajectory, 
despite the hyperparameters being optimized for SSTE. 
This improvement is even larger for the 2-bit qReLU activation in the 
right-hand plot, where
the FTP-SH qReLU even outperforms the full-precision ReLU for part of its
trajectory, and outperforms the SSTE-trained qReLU by almost $2$\%. 
Interestingly, we find that the saturated ReLU outperforms the
standard ReLU by almost a full point of accuracy. We believe that this is due to the
regularization effect caused by saturating the activation. 
This may also account for the surprisingly good performance of the FTP-SH qReLU relative
to full-precision ReLU, as hard-threshold activations also provide a strong regularization effect.

Finally, we ran a single experiment with ResNet-18 on ImageNet, 
using hyperparameters from previous works that used SSTE, 
to check (i) whether the soft hinge loss exhibits
vanishing gradient behavior due to its diminishing slope away from the origin, and
(ii) to evaluate the performance of FTP-SH for a less-quantized ReLU
(we used $k=5$ steps, which is less than the full range of a 3-bit ReLU).
While FTP-SH does slightly worse than SSTE for the sign function, 
we believe that this is 
because the hyperparameters were tuned for SSTE and not due to vanishing gradients,
as we would expect much worse accuracy in that case. 
Results from the qReLU activation provide further evidence against vanishing
gradients as FTP-SH for qReLU outperforms SSTE by almost $4$\% in top-1 accuracy
(Table~\ref{table:accuracies}).
\looseness=-1

\vspace{-1mm}
\section{Conclusion}
\vspace{-1mm}

In this work, we presented a novel mixed convex-combinatorial optimization framework for
learning deep neural networks with hard-threshold units.
Combinatorial optimization is used to set discrete targets for the 
hard-threshold hidden units, such that each unit only has a linearly-separable
problem to solve. The network then decomposes into individual perceptrons, which
can be learned with standard convex approaches, given these targets.
Based on this, we developed a recursive algorithm for learning
deep hard-threshold networks, which we call 
feasible target propagation (\algname{}), and an efficient 
mini-batch variant (\algnamemb{}). 
We showed that the commonly-used but poorly-justified
saturating straight-through estimator (STE) 
is the special case of \algnamemb{} 
that results from using a saturated hinge
loss at each layer and our target heuristic and other types of STE correspond
to other heuristic and loss combinations in \algnamemb{}.
Finally, we defined the soft hinge loss and showed
that \algnamemb{} with a soft hinge loss 
at each layer improves classification accuracy
for multiple models on CIFAR-10 and ImageNet when compared 
to the saturating STE.


In future work, we plan to develop novel target heuristics and layer loss functions
by investigating connections between our framework and constraint satisfaction and
satisfiability. We also intend to further explore the benefits of deep networks
with hard-threshold units. 
In particular, while recent research clearly shows their ability to 
reduce computation and energy requirements, they 
should also be less susceptible to vanishing and exploding gradients
and may be less susceptible to covariate shift and adversarial examples.
\looseness=-1

\subsubsection*{Acknowledgments}
This research was partly funded by ONR grant N00014-16-1-2697. The GPU
machine used for this research was donated by NVIDIA.

{\small{
\bibliography{library}
\bibliographystyle{iclr2018_conference}
}}

\newpage
\appendix

\section{Experiment details}
\label{apx:exp}

All experiments were performed using PyTorch 
({\small \url{http://pytorch.org/}}).
CIFAR-10 experiments with the 4-layer convolutional network were
performed on an NVIDIA Titan X. 
All other experiments were performed on NVIDIA Tesla P100 devices
in a DGX-1.
Code for the experiments is available at {\small \url{https://github.com/afriesen/ftprop}}.

\subsection{CIFAR-10}

On CIFAR-10, which has $50$K training images and $10$K test images divided into $10$ classes, 
we trained both a simple $4$-layer convolutional network 
and a deeper $8$-layer convolutional network used in~\citep{Zhou2016} with the above methods 
and then compared their top-1 accuracies on the test set. 
We pre-processed the images with mean / std normalization,
and augmented the dataset with random horizontal flips and random crops from images
padded with $4$ pixels.
Hyperparameters were chosen based on a small amount of exploration on a validation set.

The first network we tested on CIFAR-10 was a simple 4-layer convolutional network (convnet)
structured as: 
$\text{conv}(32) 
\rightarrow \text{conv}(64) 
\rightarrow \text{fc}(1024) 
\rightarrow \text{fc}(10)$, 
where $\text{conv}(c)$ and $\text{fc}(c)$ indicate a convolutional layer and fully-connected layer, respectively, 
with $c$ channels. Both convolutional layers used $5 \times 5$ kernels. 
Max-pooling with stride $2$ was used after each convolutional layer, 
and a non-linearity was placed before each of the above layers except the first.
Adam~\citep{Kingma2015} with learning rate $2.5$e-$4$ 
and weight decay $5$e-$4$ was used
to minimize the cross-entropy loss for $300$ epochs.
The learning rate was decayed by a factor of $0.1$ after $200$ and $250$ epochs. 

In order to evaluate the performance of \algnamemb{} with the soft hinge loss on a deeper network, 
we adapted the 8-layer convnet from \citet{Zhou2016} to CIFAR-10. 
This network has $7$ convolutional layers and one fully-connected layer
for the output and uses batch normalization~\citep{Ioffe2015} before each non-linearity.
We optimized the cross-entropy loss with Adam using a learning rate of $1$e-$3$ and a weight decay of $1$e-$7$
for the sign activation and $5$e-$4$ for the qReLU and baseline activations.
We trained for $300$ epochs, decaying the learning rate by $0.1$ after $200$ and $250$ epochs.

\subsection{Learning curves for CIFAR-10}


\begin{figure}[h!]
  \centering
  \begin{subfigure}[b]{0.49\columnwidth}
    \includegraphics[width=\textwidth]{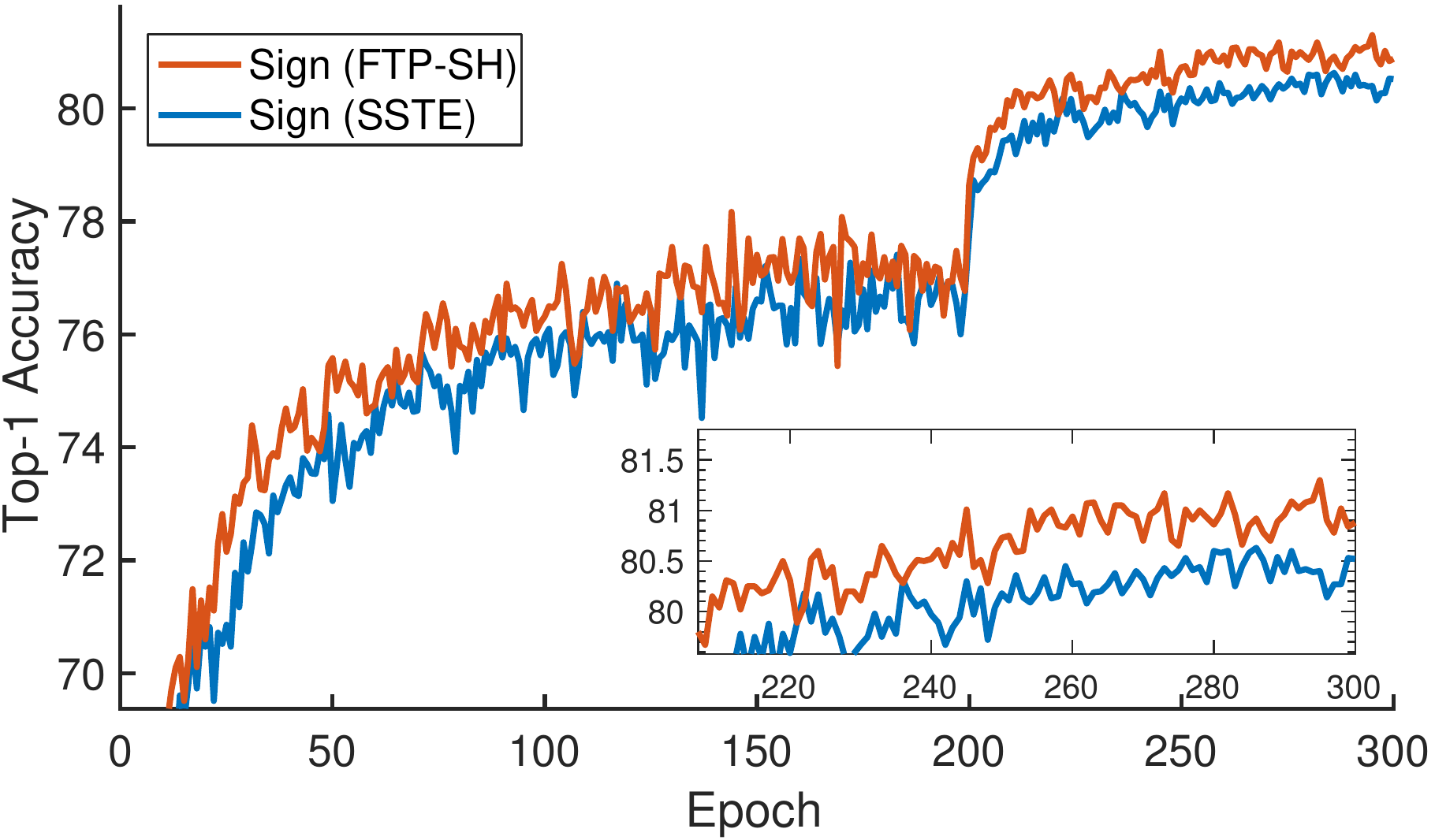}
  \end{subfigure}
  \begin{subfigure}[b]{0.49\columnwidth}
    \includegraphics[width=\textwidth]{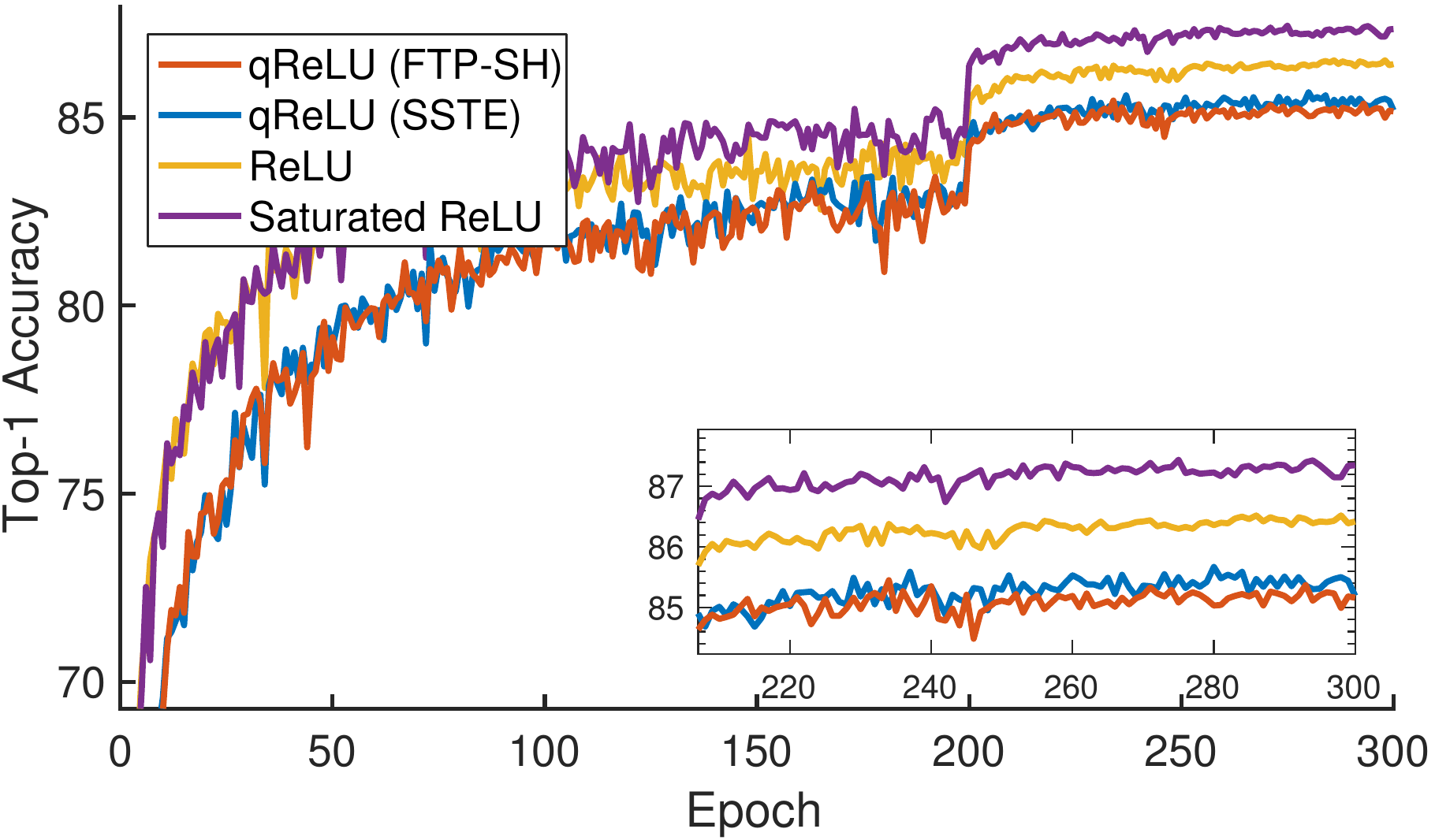}
  \end{subfigure}
  \hspace{0.05em}
  \caption{	
  \small{
The top-1 test accuracies for 
the $4$-layer convolutional network with
different activation functions on CIFAR-10. 
The inset figures show the test accuracy for the final $100$ epochs in detail. 
The left figure shows the network with sign activations.
The right figure shows the network with 2-bit quantized ReLU (qReLU) activations
and with the full-precision baselines.
Best viewed in color.}
}
\label{fig:cifar10_conv4} 
\end{figure}

\begin{figure}[H]
  \centering
  \begin{subfigure}[b]{0.49\columnwidth}
    \includegraphics[width=\textwidth]{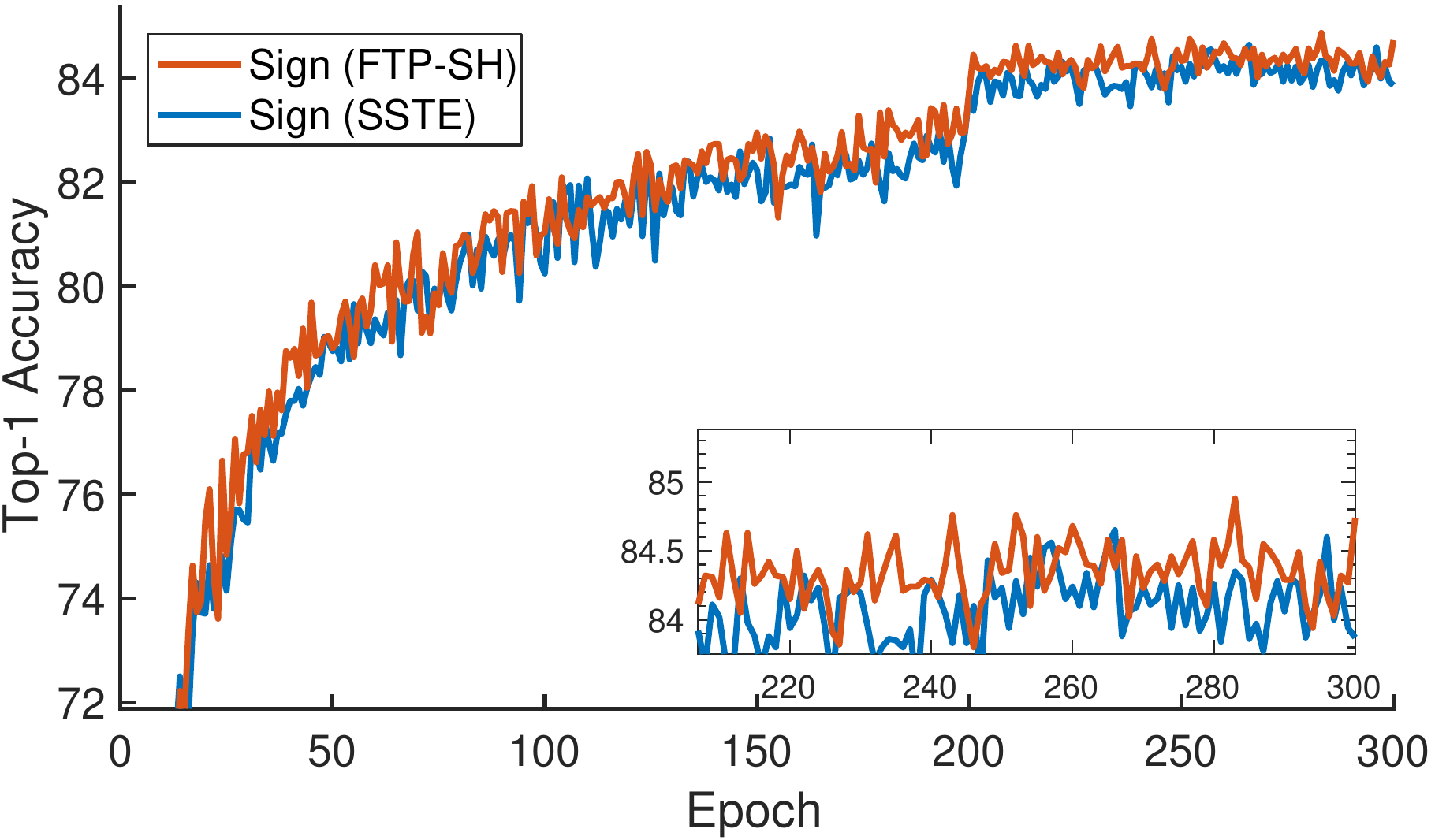}
  \end{subfigure}
  \begin{subfigure}[b]{0.49\columnwidth}
    \includegraphics[width=\textwidth]{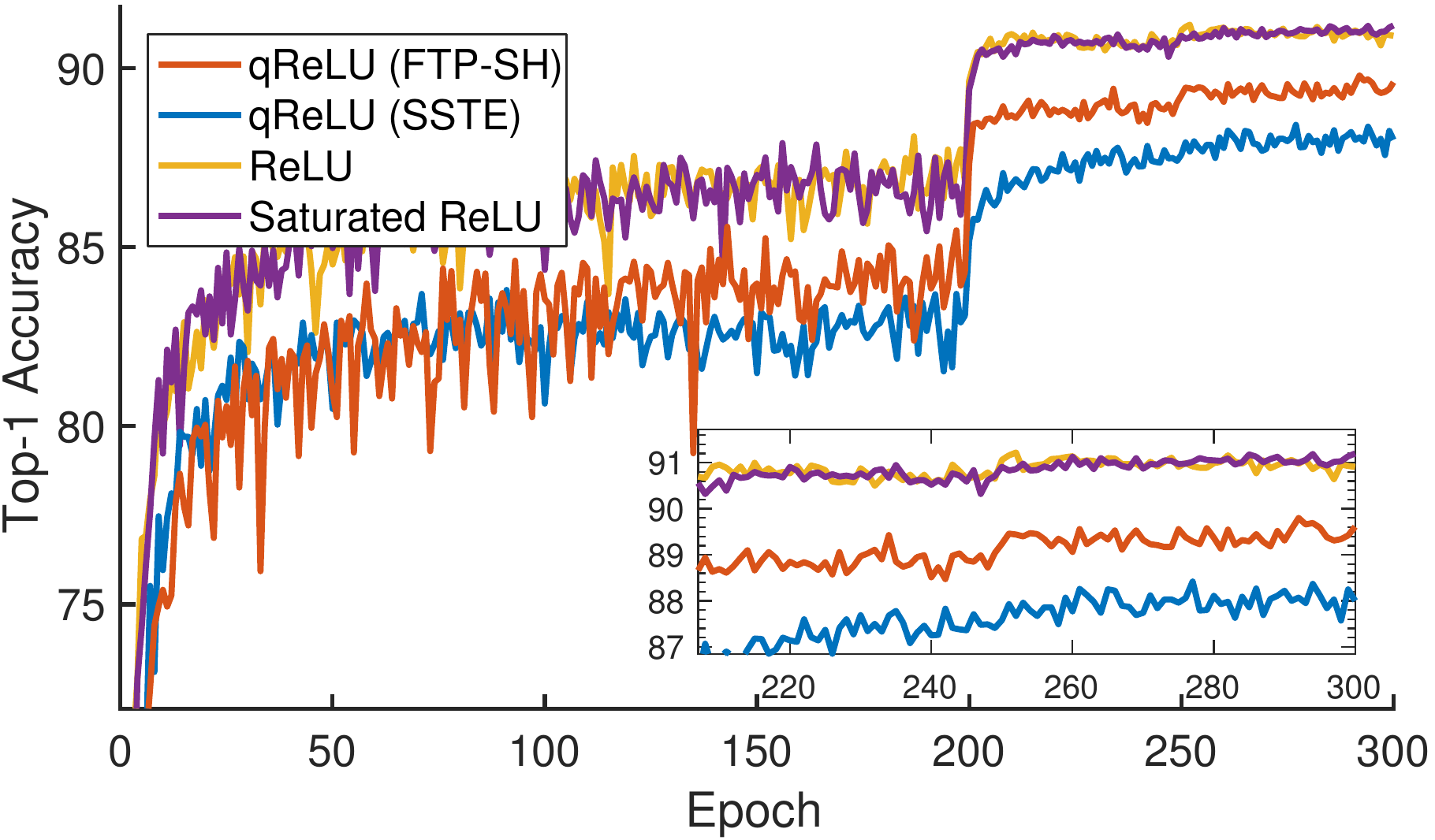}
  \end{subfigure}
  \hspace{0.05em}
  \caption{	
  \small{
The top-1 test accuracies for 
the $8$-layer convolutional network with
different activation functions on CIFAR-10. 
The inset figures show the test accuracy for the final $100$ epochs in detail. 
The left figure shows the network with sign activations.
The right figure shows the network with 2-bit quantized ReLU (qReLU) activations
and with the full-precision baselines.
Best viewed in color.}
}
\label{fig:cifar10_conv8} 
\end{figure}

~\\

\subsection{ImageNet (ILSVRC 2012)}

On ImageNet, a much more challenging dataset with roughly $1.2$M training images and 
$50$K validation images divided into $1000$ classes, 
we trained AlexNet, the most commonly used model in the quantization literature, 
with different activations and compared top-1 and top-5 accuracies of the trained models 
on the validation set.  As is standard practice, we treat the validation set as the test data.
Images were resized to $256 \times 256$, mean / std normalized, and then 
randomly cropped to $224 \times 224$ and randomly horizontally flipped.
Models are tested on centered $224 \times 224$ crops of the test images.
Hyperparameters were set based on \citet{Zhou2016} and \citet{Zhu2017}, 
which both used SSTE to train AlexNet on ImageNet.

We trained the \citet{Zhou2016} variant of AlexNet~\citep{Krizhevsky2012}
on ImageNet with sign, 2-bit qReLU, ReLU, and saturated ReLU activations.
This version of AlexNet removes the dropout and replaces the local contrast normalization
layers with batch normalization. Our implementation does not split the convolutions
into two separate blocks. We used the Adam optimizer with
learning rate $1$e-$4$ on the cross-entropy loss for $80$ epochs, decaying the
learning rate by $0.1$ after $56$ and $64$ epochs.
For the sign activation, we used a weight decay of $5$e-$6$ as in \citet{Zhou2016}.
For the ReLU and saturated ReLU activations, which are much more likely
to overfit, we used a weight decay of $5$e-$4$, as used in \citet{Krizhevsky2012}.
For the 2-bit qReLU activation, we used a weight decay of $5$e-$5$, since
it is more expressive than sign but less so than ReLU.

As with AlexNet, we trained ResNet-18~\citep{He2015a} on ImageNet with
sign, qReLU, ReLU, and saturated ReLU activations; however, for ResNet-18
we used a qReLU with $k=5$ steps (i.e., 6 quantization levels, 
requiring $3$ bits).
We used the ResNet code provided by PyTorch.
We optimized the cross-entropy loss with SGD with learning rate $0.1$ and momentum $0.9$
for $90$ epochs, decaying the learning rate by a factor of $0.1$ after $30$ and $60$
epochs. For the sign activation, we used a weight decay of $5$e$-7$. 
For the ReLU and saturated ReLU activations, we used a weight decay of $1$e-$4$.
For the qReLU activation, we used a weight decay of $1$e-$5$.

\newpage
\subsection{Learning curves for ImageNet}

\begin{figure}[htb]
  \centering
  \begin{subfigure}[b]{0.49\columnwidth}
    \includegraphics[width=\textwidth]{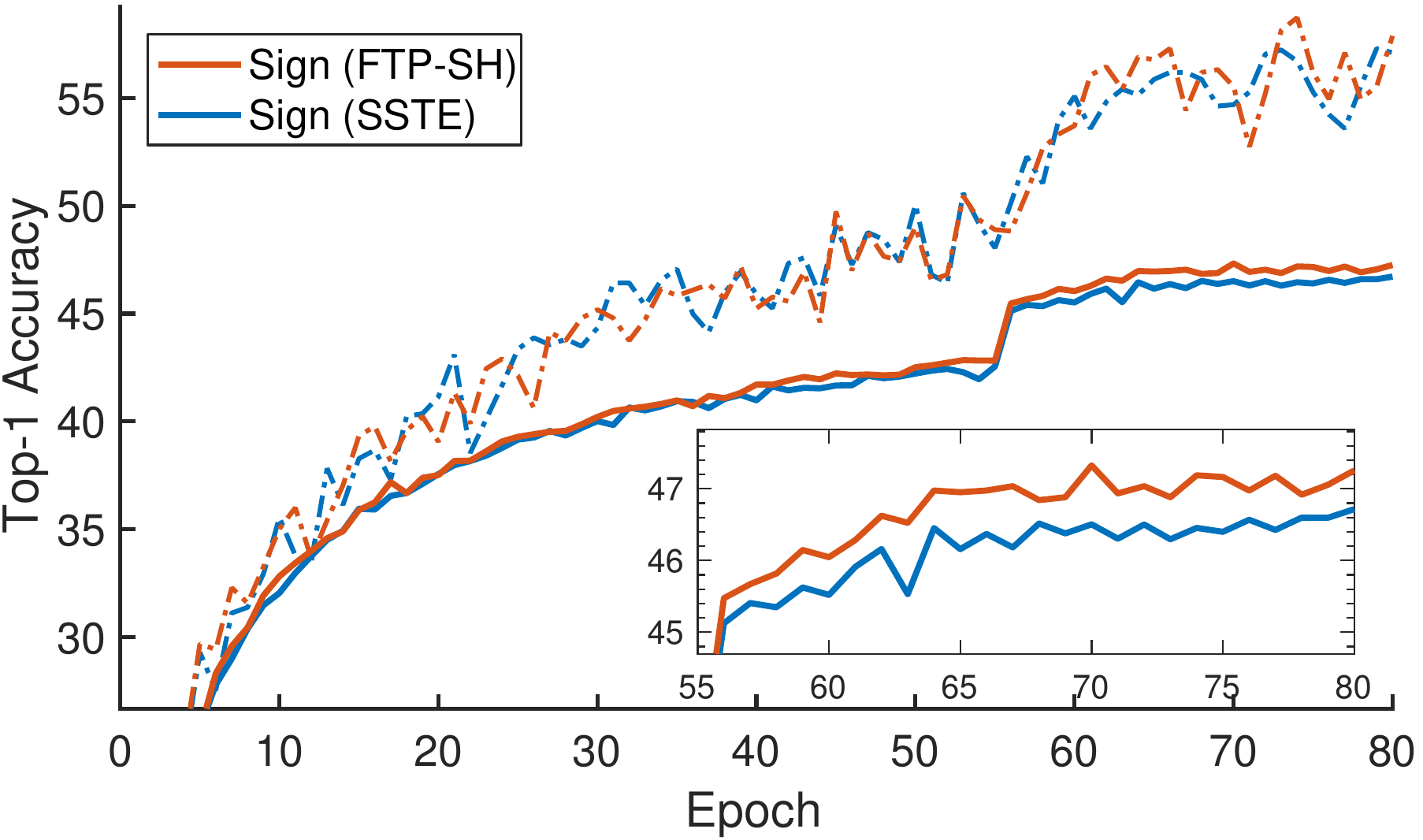}
  \end{subfigure}
  \begin{subfigure}[b]{0.49\columnwidth}
    \includegraphics[width=\textwidth]{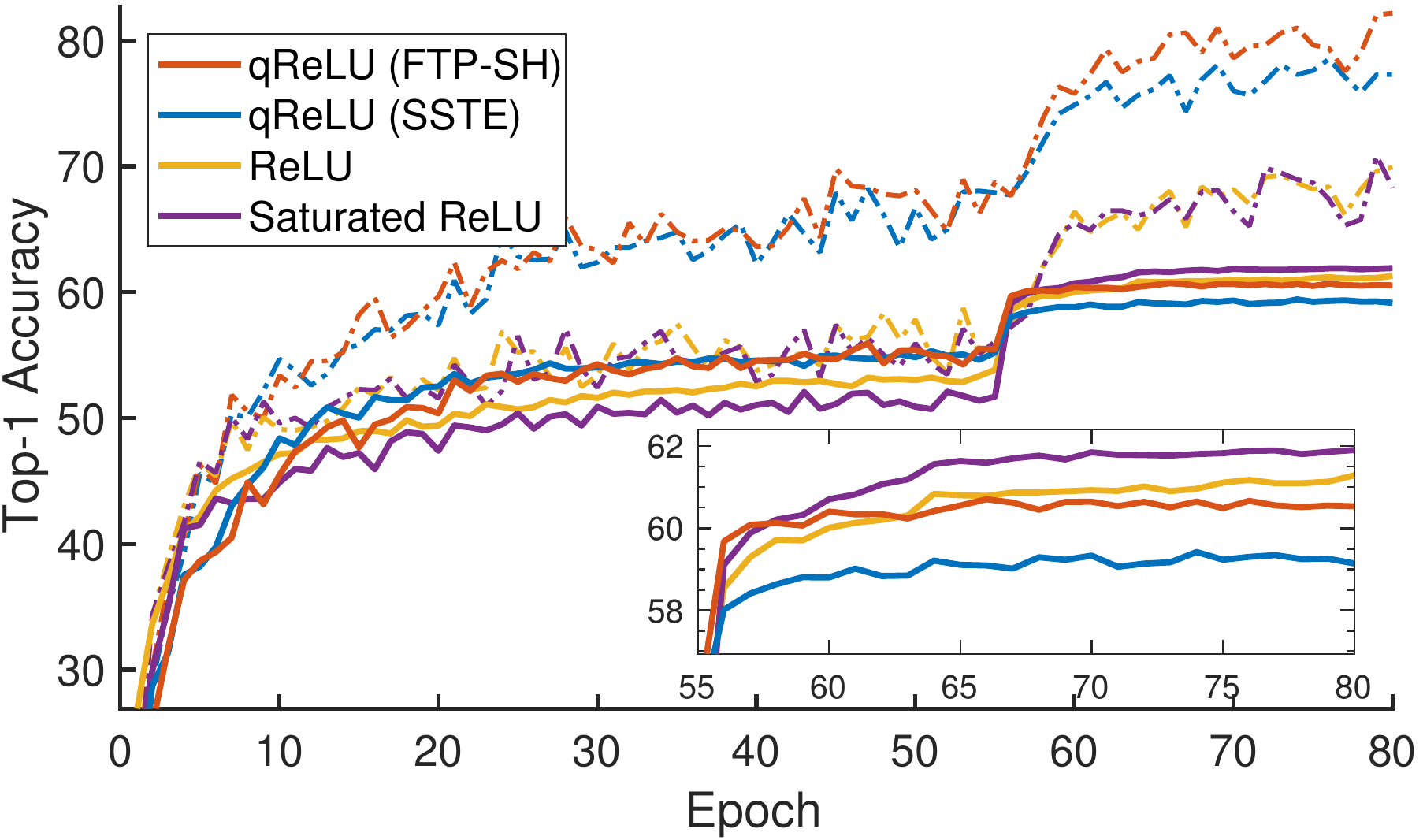}
  \end{subfigure}
  \hspace{0.05em}
  \caption{	
  \small{
The top-1 train (thin dashed lines) and test (thicker solid lines) 
accuracies for AlexNet with
different activation functions on ImageNet. 
The inset figures show the test accuracy for the final $25$ epochs in detail. 
The left figure shows the network with sign activations.
The right figure shows the network with 2-bit quantized ReLU (qReLU) activations
and with the full-precision baselines.
Best viewed in color.}
}
\label{fig:imagenet_alexnet_appendix} 
\end{figure}

\begin{figure}[htb]
  \centering
  \begin{subfigure}[b]{0.49\columnwidth}
    \includegraphics[width=\textwidth]{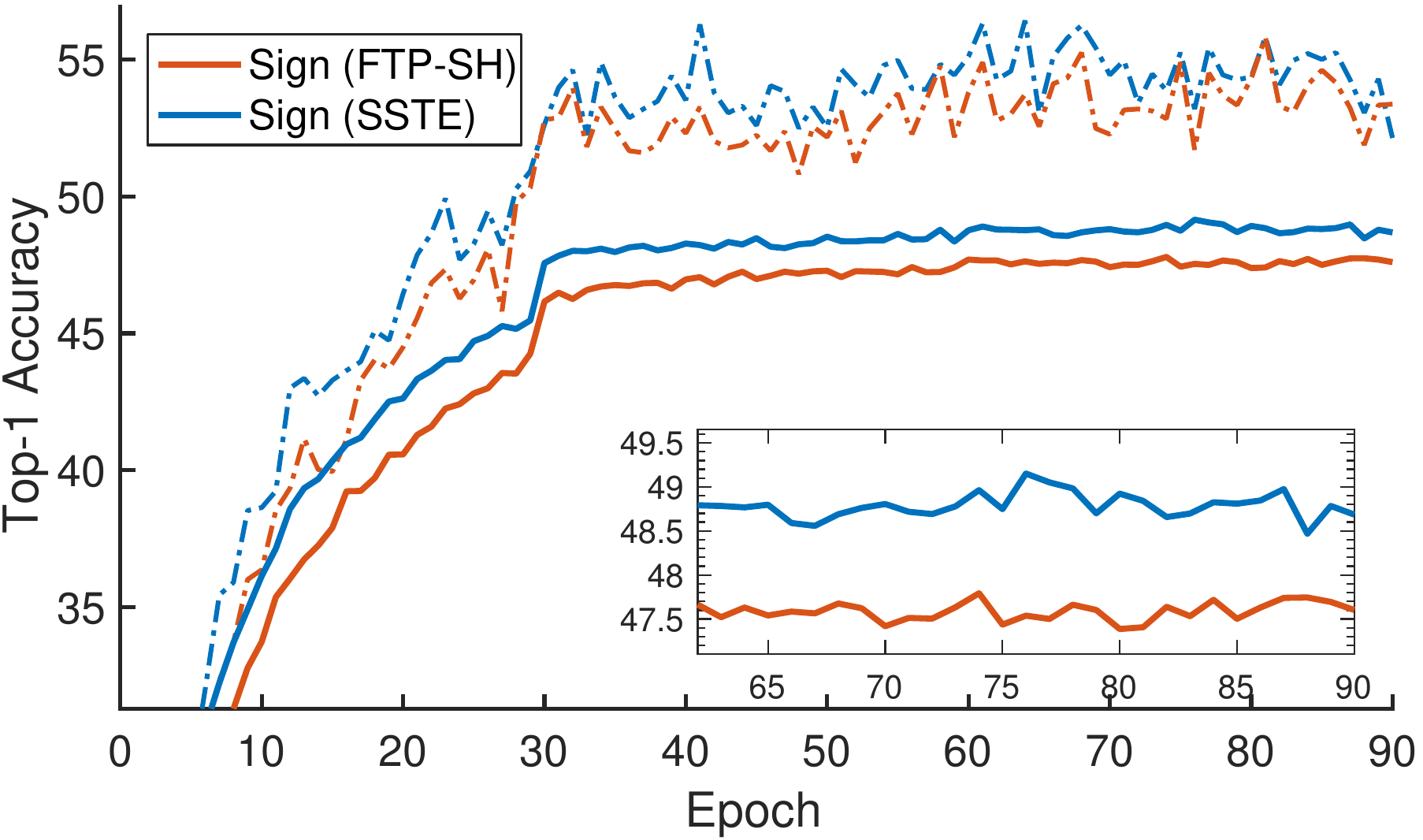}
  \end{subfigure}
  \begin{subfigure}[b]{0.49\columnwidth}
    \includegraphics[width=\textwidth]{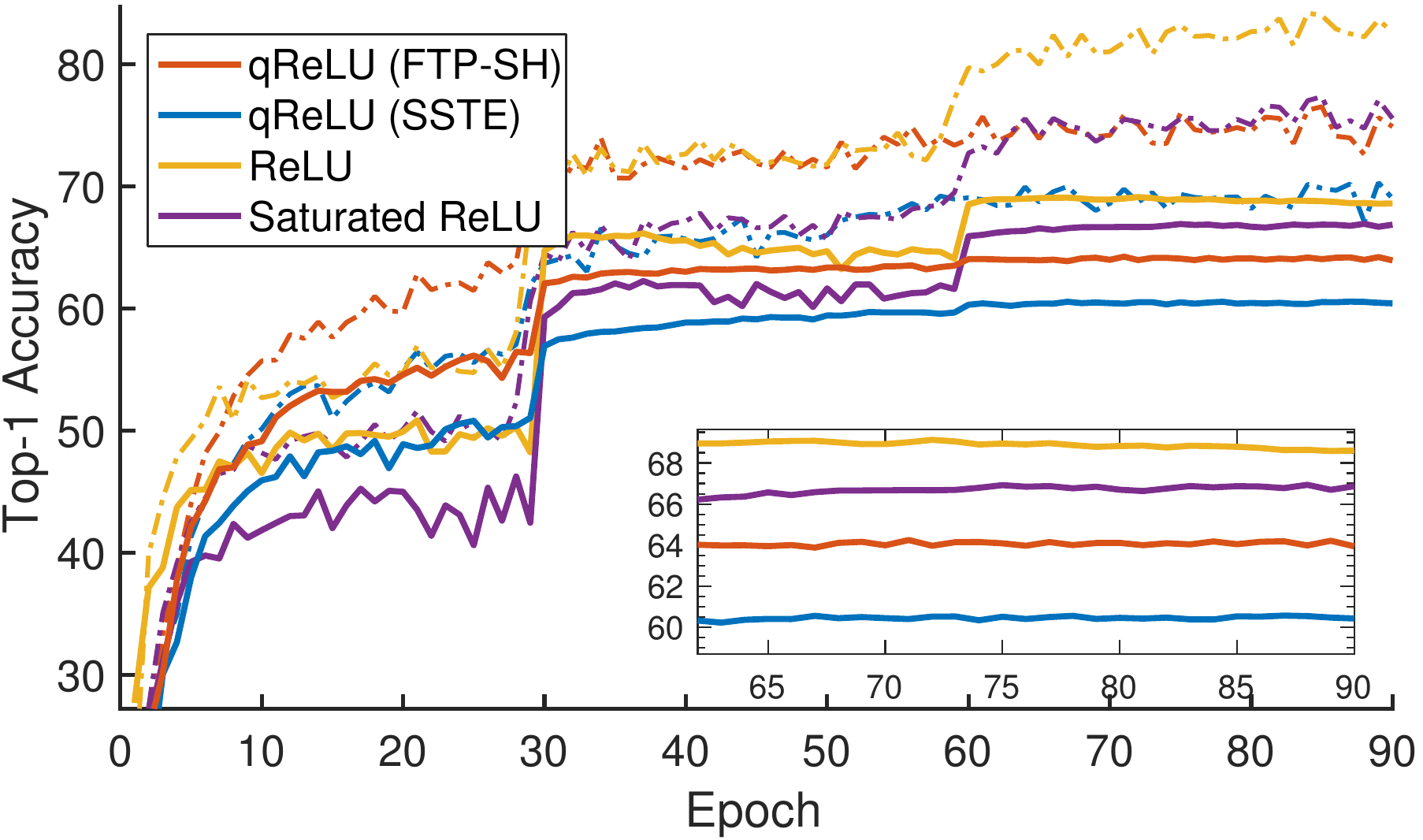}
  \end{subfigure}
  \hspace{0.05em}
  \caption{	
  \small{
The top-1 train (thin dashed lines) and test (thicker solid lines) accuracies for ResNet-18 with
different activation functions on ImageNet. 
The inset figures show the test accuracy for the final $60$ epochs in detail. 
The left figure shows the network with sign activations.
The right figure shows the network with 3-bit quantized ReLU (qReLU) activations
and with the full-precision baselines.
Best viewed in color.}
}
\label{fig:imagenet_resnet} 
\end{figure}

\end{document}